%% file: main-arxiv.tex
\pdfoutput=1

\documentclass[11pt]{article}
\usepackage[]{acl}
\usepackage{times}
\usepackage{latexsym}
\usepackage{comment}
\usepackage[T1]{fontenc}
\usepackage[utf8]{inputenc}
\usepackage{microtype}
\usepackage[utf8]{inputenc}   
\usepackage[T1]{fontenc}      
\usepackage{hyperref}         
\usepackage{url}              
\usepackage{booktabs}         
\usepackage{amsfonts}         
\usepackage{nicefrac}         
\usepackage{microtype}        
\usepackage[most]{tcolorbox}   
\usepackage{xcolor}   
\input{preamble}

\usepackage{subfigure}
\usepackage{graphbox}
\usepackage{svg}
\usepackage{nicefrac}
\usepackage{graphicx}
\definecolor{thedarkblue}{RGB}{0,0,120}   
\definecolor{mydarkblue}{rgb}{0,0.08,0.45}   

\usepackage{hyperref}
\hypersetup{  
    colorlinks=true,
    linkcolor=mydarkblue,
    citecolor=mydarkblue,
    filecolor=mydarkblue,
    urlcolor=mydarkblue}

\usepackage{subfigure}
\usepackage{nicefrac}

\DeclareMathAlphabet{\mathbcal}{OMS}{cmsy}{b}{n}
\usepackage{amsmath}
\usepackage{mathrsfs}
\usepackage{comment}

\usepackage{bm}
\usepackage{bbm}
\usepackage{amssymb}
\usepackage{enumitem}
\usepackage{paralist}

\usepackage{wrapfig,lipsum,booktabs}

\usepackage{adjustbox}
\usepackage{array}

\newcolumntype{R}[2]{  
    >{\adjustbox{angle=#1,lap=\width-(#2)}\bgroup}  
    l  
    <{\egroup}  
}
\newcommand*\rot{\multicolumn{1}{R{90}{1em}}}  

\title{A Survey of Small Language Models}

\author{  
{\bf Chien Van Nguyen$^1$\thanks{*The authors contributed equally to this work.}\;,
Xuan Shen$^{2*}$,
Ryan Aponte$^{3*}$,
Yu Xia$^4$,}
Samyadeep Basu$^5$, \\
{\bf Zhengmian Hu$^5$,
Jian Chen$^6$,
Mihir Parmar$^7$,
Sasidhar Kunapuli,
Joe Barrow$^8$,} \\
{\bf Junda Wu$^4$, Ashish Singh$^9$,
Yu Wang$^1$,
Jiuxiang Gu$^8$,
Franck Dernoncourt$^8$,} \\
{\bf Nesreen K. Ahmed$^{10}$,
Nedim Lipka$^8$,
Ruiyi Zhang$^8$,
Xiang Chen$^8$,
Tong Yu$^8$,} \\
{\bf Sungchul Kim$^8$, 
Hanieh Deilamsalehy$^8$,
Namyong Park$^{11}$,
Mike Rimer,
Zhehao Zhang$^{12}$,} \\
{\bf Huanrui Yang$^{13}$,
Ryan A. Rossi$^8$,
Thien Huu Nguyen$^1$}\\ \\ 
$^1$University of Oregon, $^2$Northeastern University, $^3$Carnegie Mellon University \\
$^4$University of California, San Diego, $^5$University of Maryland, College Park \\
$^6$State University of New York at Buffalo, $^7$Arizona State University \\
$^8$Adobe Research, $^9$University of Massachusetts Amherst, $^{10}$Intel AI Research \\
$^{11}$Meta AI, $^{12}$Dartmouth College, $^{13}$University of Arizona \\
}
\begin{document}

\maketitle

\begin{abstract}
Small Language Models (SLMs) have become increasingly important due to their efficiency and performance to perform various language tasks with minimal computational resources, making them ideal for various settings including on-device, mobile, edge devices, among many others.
In this article, we present a comprehensive survey on SLMs, focusing on their architectures, training techniques, and model compression techniques.
  
We propose a novel taxonomy for categorizing the methods used to optimize SLMs, including model compression, pruning, and quantization techniques.
We summarize the benchmark datasets that are useful for benchmarking SLMs along with the evaluation metrics commonly used.
Additionally, we highlight key open challenges that remain to be addressed.
  
Our survey aims to serve as a valuable resource for researchers and practitioners interested in developing and deploying small yet efficient language models.
  
\end{abstract}

\section{Introduction}

Although large language models (LLMs) have demonstrated impressive performance on a wide array of benchmarks and real-world situations, their success comes at significant cost.
LLMs are resource-intensive to train and run, requiring significant compute \textit{and} data. This often means that they are run on centralized and specialized hardware for both training and inference.

As a response to these challenges, there has been a growing interest in small language models (SLMs).
Small language models aim to retain the accuracy and/or adaptability of large language models, while being subject to some constraint(s), such as training or inference hardware, data availability, bandwidth, or generation time.
Improving model performance relative to these constraints can then improve downstream goals such as privacy, cost, or the ability to run on consumer devices.

The inherent difficulty of a survey of small language models is that the definitions of ``small'' and ``large'' are a function of both context and time. GPT-2, a ``large language model'' in 2019 at 1.5B parameters, is smaller than many ``small'' language models covered in this survey.
However, although the scale changes, the goals of training small language models remain relatively stable.

In this survey, we explore the architectures, training, and model compression techniques that enable the building and inferencing of SLMs.
In addition, we summarize the benchmark datasets and evaluation metrics commonly used in evaluating SLM performance.
To do this, we propose a novel taxonomy for organizing the methods along two axes:
\begin{compactitem}
\item the \textbf{techniques} used in pre-processing (model architecture), training, and post-processing (model compression) SLMs; and
\item the \textbf{constraints} the technique is attempting to optimize for, e.g. inference compute, training time, speed, etc.
\end{compactitem}
An overview of these axes can be found in Table~\ref{table:techniques} (techniques) and Table~\ref{table:settings} (constraints).

It is important to note that progress on any one of these goals does not necessarily imply progress on the others. In fact, there are often trade-offs between them. For instance, memory-efficient training methods like quantization-aware training ~\citep{dettmers2022gpt3,dettmers2024qlora} are often slower than their full-precision counterparts. However, by using mixed precision to represent the weights and gradients, they allow training or finetuning using less memory.
Finally, although there have been several recent surveys on LLMs and their learning methods \cite{rogers-etal-2020-primer, Min2021RecentAI, Zhu2023ASO, Shen2023LargeLM}, to the best of our knowledge, this is the first survey focused on SLMs.

\begin{table*}[ht!]
\centering
\small
\begin{tabular}{l r cccccc}
\toprule
\textbf{Technique} & \textbf{General Mechanism} & \rot{\textbf{Training Compute}} &\rot{\textbf{Dataset Size}}  &\rot{\textbf{Inference Runtime}} & \rot{\textbf{Memory}} & \rot{\textbf{Storage Space}} & \rot{\textbf{Latency}} \\

\midrule
\multirow{3}{*}{\textbf{Model Architectures} (Sec. \ref{sec:model-architecture})} 
  
    & Lightweight Models (Sec.~\ref{sec:lightweight-arch}) & \checkmark & & \checkmark & \checkmark & & \checkmark  \\
    \cmidrule{2-8 }
  
& Efficient Self-Attention (Sec.~\ref{sec:model-comp-low-rank-factorization}) & \checkmark & & \checkmark & \checkmark & & \checkmark \\
    \cmidrule{2-8}
    & Neural Arch. Search (Sec.~\ref{sec:nas-techniques}) & & & \checkmark &\checkmark & \checkmark &  \\
\midrule
\multirow{3}{*}{\textbf{Training Techniques} (Sec. \ref{sec:training-techniques})} 
  
    & Pre-training (Sec. \ref{sec:pre-training}) & \checkmark & \checkmark & \checkmark & \checkmark & \checkmark &  \\
    \cmidrule{2-8}
    & Finetuning (Sec.~\ref{sec:Fine-tuning-Technique}) & \checkmark & \checkmark & & & & \\
\midrule
\multirow{4}{*}{\textbf{Model Compression} (Sec. \ref{sec:model-compression-techniques})} 
  
    & Pruning (Sec.~\ref{sec:model-comp-pruning-techniques}) & & & \checkmark & \checkmark & \checkmark & \checkmark \\   
    \cmidrule{2-8}
      
    & Quantization (Sec.~\ref{sec:model-compression-quant}) & & & \checkmark & \checkmark & \checkmark & \checkmark  \\
    \cmidrule{2-8}
    & Knowledge Distillation (Sec.~\ref{sec:knowledge-distil-tech}) & & \checkmark & & & & \\

\bottomrule
\end{tabular}
\caption{General techniques used for optimizing small language models, categorized by type of model optimization and most central constraints they address. 
}
\label{table:techniques}
\end{table*}

\medskip\noindent\textbf{Organization of the Survey.} 
This survey is structured into three main sections, each covering a key aspect of optimizing SLMs.
\textbf{Section~\ref{sec:model-architecture}} focuses on model architectures, including lightweight designs, efficient self-attention approximations, and neural architecture search to efficiently build smaller models.
\textbf{Section~\ref{sec:training-techniques}} covers efficient pre-training and fine-tuning techniques to enhance performance for SLMs while managing resource constraints.
\textbf{Section~\ref{sec:model-compression-techniques}} explores model compression techniques, such as pruning, quantization, and knowledge distillation, which reduce model size and latency without sacrificing significant accuracy.
\textbf{Section~\ref{sec:data-and-evaluation}} introduces an overview of benchmark datasets and evaluation metrics, providing a comprehensive framework for assessing the effectiveness of these methods.
\textbf{Section~\ref{sec:applications}} discusses the applications that are enabled by SLMs, organized by constraints.
Finally, a discussion of open challenges for SMLs is presented in \textbf{Section~\ref{sec:open-problems-challenges}}.

\noindent\textbf{Summary of Main Contributions.} 
The key contributions of this work are as follows:

\begin{compactitem}
    \item A comprehensive survey of existing work on small language models for practitioners.
    We also survey the problem settings, evaluation metrics, and datasets used in the literature.

    \item We introduce a few intuitive taxonomies for SLMs and survey existing work using these taxonomies.

    \item We identify important applications, open problems, and challenges of SLMs for future work to address. 
\end{compactitem}

\section{Model Architectures} \label{sec:model-architecture}
This section discusses the architectural designs for developing SLMs. Specifically, we cover lightweight architectures (Section \ref{sec:lightweight-arch}), 
  
efficient self-attention approximations (Section~\ref{sec:model-comp-low-rank-factorization}),
and neural architecture search (Section \ref{sec:nas-techniques}).
  
\subsection{Lightweight Architectures} \label{sec:lightweight-arch}

Lightweight language model architectures are designed to achieve efficient performance with fewer parameters and reduced computational overhead, 
which is ideal for deployment on resource-constrained devices such as mobile phones, edge devices, and embedded systems.
Representative lightweight models often follow the encoder-only and decoder-only architectures.

Lightweight encoder-only architectures are mostly optimized versions of BERT \cite{devlin-etal-2019-bert}.
For example, MobileBERT \cite{sun-etal-2020-mobilebert} introduces an inverted-bottleneck structure to maintain a balance between self-attention and feed-forward networks, achieving a 4.3x size reduction and a 5.5x speedup compared to the base version of BERT. DistilBERT \cite{sanh2019distilbert} and TinyBERT \cite{jiao2019tinybert} achieve more than 96\    

Lightweight decoder-only architectures follow the structure of autoregressive language models such as the GPT \cite{Radford2018Improving,radford2019language} and LLaMA series \cite{touvron2023llamaopenefficientfoundation}.
These models emphasize knowledge distillation, memory overhead optimization, parameter sharing, embedding sharing to enhance efficiency and scalability. 
BabyLLaMA~\cite{timiryasov2023baby} and BabyLLaMA-2~\cite{tastet2024babyllama} distill knowledge from multiple teachers into a 58M-parameter model and a 345M-parameter model respectively, demonstrating that distillation can exceed teacher models' performance particularly under data-constrained conditions. TinyLLaMA~\cite{zhang2024tinyllama}, with only 1.1B parameters, achieves high efficiency by optimizing memory overhead, e.g., via FlashAttention \cite{dao2022flashattention}, while maintaining competitive performance for various downstream tasks.
MobilLLaMA~\cite{thawakar2024mobillama} applies a parameter-sharing scheme that reduces both pre-training and deployment costs, introducing a 0.5B-parameter model for resource-constrained devices.
MobileLLM~\cite{liu2024mobilellm} further introduces embedding-sharing and grouped-query attention mechanisms with block-wise weight sharing to reduce latency.

\subsection{Efficient Self-Attention Approximations}
\label{sec:model-comp-low-rank-factorization}
 
Deploying large language models can be challenging due to the substantial number of parameters in the self-attention layers, as well as the computational cost associated with self-attention. In this section, we discuss strategies towards decreasing this computational cost which can ultimately be useful in creating small language models.
  
Reformer~\cite{kitaev2020reformer} improves the complexity of the self-attention from $\mathcal{O}(N^{2})$ to $\mathcal{O}(N \log N)$ by replacing the dot product attention with one which uses locality-sensitivity hashing. ~\citet{roy2021efficient} use a sparse routing module based on an online k-means clustering, which reduces the complexity of the attention computation. 

To reduce the computational quadratic complexity of the self-attention layer from $\mathcal{O}(N^{2})$ to $\mathcal{O}(N)$, several works, including \cite{wang2020linformer, katharopoulos2020transformers, xiong2021nystromformer, DBLP:journals/corr/abs-2004-05150}, propose linear attention mechanisms. In particular,~\cite{katharopoulos2020transformers} express self-attention as a linear dot-product of kernel feature maps, thus reducing the quadratic complexity. The authors further show that transformers with this linear attention mechanism can be viewed as a recurrent neural network which enables faster inference. Building on these foundations, recent advancements have led to more advanced architectures. Notable examples include Mamba \cite{gu2023mamba, pmlr-v235-dao24a}, which introduces a selective state space model with input-dependent transitions, and RWKV \cite{peng-etal-2023-rwkv}, which combines elements of transformers and RNNs with a linear attention mechanism. These models not only achieve linear time and space complexity but also demonstrate competitive performance across various tasks. This ongoing trend towards efficient sequence modeling architectures aims to maintain the expressiveness of attention-based models while significantly reducing computational complexity.

We also note some previous work for processing long documents with encoder-only architectures. Longformer~\cite{DBLP:journals/corr/abs-2004-05150} uses a combination of local windowed attention and task-specific global attention which scales linearly with input length, thus being memory efficient. 
~\citet{wang2020linformer} approximates the self-attention mechanism using a low-rank matrix which reduces the complexity to $\mathcal{O}(N)$. Both these works show that empirically transformers with linear self-attention matches the performance of the original self-attention mechanism across a variety of downstream tasks. In a similar vein,~\citet{xiong2021nystromformer} use the popular Nystrom method~\cite{10.1007/BF02547521} for approximating the self-attention operation with strong empirical performances when compared to traditional transformers.

\subsection{Neural Architecture Search Techniques} \label{sec:nas-techniques}

This section discusses automated methods to discover the most efficient model architectures for specific tasks and hardware constraints.

Previous research has primarily concentrated on Neural Architecture Search (NAS) for vision tasks~\cite{tan2019efficientnet, zoph2016neural, wu2019fbnet, guo2020single} and BERT models~\cite{xu2021bert, jawahar2023mixture, ganesan2021supershaper}, as these models have comparatively fewer parameters, which reduces the cost of the search process for efficient architectures. However, LLMs with over a billion parameters present a significant challenge in searching for smaller, more efficient models.
Their massive scale makes the search process computationally intensive and costly.
Recently, MobileLLM~\cite{liu2024mobilellm} investigates the impact of model depth (i.e., number of layers) and width (i.e., number of heads) on performance, effectively conducting a targeted architecture search within a smaller parameter range for language models with millions of parameters.
Meanwhile, \citet{shen2024search} reduce the search space by exploring an appropriate initialization for the search, which helps expedite the convergence of the search process.

\subsection{Small Multi-modal Models}
\label{sec:multimodel}
Recent large multi-modal models (LMMs) have achieved comparable or superior performance to their predecessors while significantly reducing the number of parameters. Notable examples include the LLaVA-Next \cite{liu2024llavanext}, Idefics2 \cite{laurenccon2024matters}, and InternVL2 \cite{chen2023internvl} series. This progress is partly driven by more efficient, smaller language models like Gemma \cite{team2024gemma}, phi-3-mini \cite{abdin2024phi}, and emphasizes the critical role of curated datasets. Additionally, there has been a concerted effort to reduce the size of the vision encoder during multi-modal fusion. InternVL2, for example, leverages outputs from intermediate layers of large visual encoders while discarding the later blocks. Smaller models, such as PaliGemma \cite{beyer2024paligemma} and Mini-Gemini \cite{li2024mini}, adopt lightweight vision encoders. Monolithic multi-modal models take this further by completely eliminating the visual encoder, instead using lightweight architectures to generate visual tokens. For example, Chameleon \cite{team2024chameleon} employs a VQ-VAE model to encode and decode images into discrete tokens, while Mono-InternVL \cite{luo2024mono} uses an MLP to generate visual tokens for image patches, incorporating a modality-specific feed-forward network, termed multi-modal Mixture-of-Experts, to differentiate between modalities.

\section{Training Techniques} 
\label{sec:training-techniques}
This section reviews the key training techniques used for language model pretraining and fine-tuning. While SLMs involve similar training approaches to LLMs, we will focus on efficient techniques to facilitate the general learning scenarios with limited resources for SLMs.

\subsection{Pre-training Techniques}\label{sec:pre-training}
Mixed precision training is a crucial technique for enhancing pre-training efficiency of SLMs and LLMs. This approach leverages low-precision representations for forward and backward propagation while maintaining high-precision weights for updates. For instance, \cite{micikevicius2018mixed} introduced Automatic Mixed Precision (AMP), which initially keeps a master copy of weights in 32-bit floating-point (FP32) precision while performing arithmetic operations in 16-bit floating-point (FP16) precision. However, recent work \cite{rae2021scaling} has observed accuracy losses due to its limited numerical range. To address this issue,  \cite{burgess2019bfloat16} propose Brain Floating Point (BFLOAT16), offering a greater dynamic range with more exponent bits than FP16. BFLOAT16 has demonstrated superior training performance and representation accuracy compared to FP16. Modern GPU architectures have further advanced mixed-precision capabilities through specialized Tensor Cores. For instance, while earlier generations supported FP16 and BFLOAT16, NVIDIA's latest Hopper architecture introduces support for 8-bit floating-point (FP8) precision \cite{luo2402benchmarking}, enabling even greater computational efficiency for large-scale language models.

Complementing these mixed precision approaches, various optimization and stability techniques are employed to prevent model collapse and further enhance training efficiency for SLMs and LLMs. 
While Adam \cite{diederik2014adam} and AdamW \cite{loshchilov2018decoupled} optimizers are commonly used, memory-efficient variants like Adafactor \cite{shazeer2018adafactor} and Sophia \cite{liu2024sophia} have been introduced to improve training speed and efficiency. To further stabilize training, gradient clipping \cite{zhang2020improved} is widely used to prevent exploding gradients. Additionally, careful initialization strategies can provide a good starting point for model training. These combined techniques aim to achieve optimal training efficiency, maintain numerical stability, and produce more robust and capable language models.

To address the computational demands of the pre-training stage, language models are typically pre-trained across multiple machine nodes, leveraging distributed computing resources efficiently. Several system-level optimization techniques have been developed to this end. Zero Redundancy Data Parallelism (ZeRO) \cite{rajbhandari2020zero} offers three progressive stages of optimization, each partitioning more training states across devices: ZeRO-1 partitions optimizer states, ZeRO-2 adds gradient partitioning, and ZeRO-3 further partitions model parameters. PyTorch's Fully Sharded Data Parallel (FSDP) \cite{zhao2023pytorch} implements similar concepts. These parallelism techniques enable training with larger batch sizes, significantly improving efficiency and scalability for SLMs and LLMs.

\subsection{Fine-tuning Techniques} \label{sec:Fine-tuning-Technique}
Fine-tuning on smaller, task-specific datasets allows LLMs to leverage the knowledge gained during pre-training, enabling them to excel in specialized tasks or domains. Fine-tuning techniques are designed to address challenges like limited computing resources, data quality, availability, and robustness, ensuring efficient adaptation to new tasks without extensive retraining.

\subsubsection{Parameter-Efficient Fine-Tuning}
Parameter-Efficient Fine-Tuning (PEFT) updates a small subset of parameters or adds lightweight modules, keeping most of the pre-trained model’s parameters fixed. This approach reduces computational costs during SLM fine-tuning,
preserves the model’s knowledge,
reduces overfitting,
and improves flexibility.
LoRA uses low-rank decomposition~\cite{hu2021lora}, Prompt Tuning \cite{lester2021power} inserts learnable prompts into inputs, and Llama-Adapter \cite{zhang2023llama, gao2023llamaadapterv2} adds prompts to LLaMA’s attention blocks. Dynamic Adapters \cite{kong2024lora, feng2024mixture, gou2023mixture, liu2023moelora, luo2024moelora} automatically combine multiple adapters as a mixture-of-experts model to enable multi-tasking and prevent forgetting \cite{han2024slim, yang2024moral}.

\subsubsection{Data Augmentation} 
Data augmentation increases the complexity, diversity and quality of training data, leading to improved generalization and performance on downstream tasks.
AugGPT \cite{dai2023auggpt} rephrases training samples using ChatGPT. Evol-Instruct \cite{xu2023wizardlm} uses multistep revisions to generate diverse, open-domain instructions with increased complexity. Reflection-tuning \cite{li2023reflection, li2024selective} enhances data quality and instruction-response consistency for instruction tuning by refining both instructions and responses using GPT-4 based on predefined criteria. FANNO \cite{zhu2024fanno} augments instructions and generates responses by incorporating external knowledge sources through retrieval-augmented generation. LLM2LLM \cite{lee2024llm2llm}  generates more hard samples based on model prediction on training data during training.

Data augmentation is also effective for synthesizing new data when training data is limited, such as for low-resource languages \cite{whitehouse2023llm}, medical and clinical applications \cite{chintagunta2021medically}, and privacy-sensitive data \cite{song2024llm}, enabling models to generalize better and perform more robustly in constrained settings.

\section{Model Compression Techniques}\label{sec:model-compression-techniques}
Model compression techniques focus on reducing the size and complexity of large pre-trained language models while maintaining their performance. As a result, these methods are a key approach to deriving SLMs from LLMs. In this section, we propose a taxonomy for model compression that categorizes such techniques by whether they perform 
pruning (Section~\ref{sec:model-comp-pruning-techniques}), 
quantization (Section~\ref{sec:model-compression-quant}), or
knowledge distillation (Section~\ref{sec:knowledge-distil-tech}).

\subsection{Pruning Techniques} \label{sec:model-comp-pruning-techniques}

Weight pruning is a model optimization technique that reduces the number of parameters to enhance computational efficiency and lower memory usage, all while maintaining performance levels. We differentiate between two major approaches for pruning: unstructured pruning and structured pruning.

{\bf Unstructured pruning} removes less significant individual weights, offering fine-grained control and flexibility in reducing model size. For example, to perform irregular pruning on large language models, SparseGPT~\cite{frantar2023sparsegpt} reformulates the pruning task as a sparse regression problem, optimizing both the remaining and pruned weights using a layer-wise approximate regression solver. SparseGPT can efficiently handle large-scale models like OPT-175B and BLOOM-176B. Additionally, ~\cite{llmpruneadmm} integrates the ADMM~\cite{ADMM} algorithm for weight updates to further mitigate pruning errors. Wanda~\cite{sun2023simple} incorporates both weights and activations into consideration during pruning process, and eliminates the need of weight updates. The n:m pruning strategy \cite{zhou2021learning} brings unstructured pruning to model acceleration by pruning exactly \textit{n} weights out of every \textit{m}, balancing pruning flexibility and computational efficiency for significant speedups. NVIDIA's TensorRT leverages such sparse patterns to optimize memory access and reduce computational loads, accelerating inference on GPUs, particularly hardware like the A100. Notably, unstructured pruning often results in sparse matrices requiring specialized hardware or algorithms to maximize computational benefits \cite{frantar2023sparsegpt}.

{\bf Structured pruning}~\citep{wang-etal-2020-structured, santacroce2023matters,ma2023llm,tao2023structured,xia2024sheared,kurtic2024ziplm} aims to compress LLMs while maintaining performance by removing groups of parameters in a structured manner, which enables more efficient hardware implementation. A major direction in this approach concerns the sparsity of neurons in the model. For instance, \citet{li2023the} observes prevalent sparsity in feed-forward networks. \citet{liu2023deja} proposes using small neural networks for dynamic pruning based on input, termed ``contextual sparsity''. \citet{mirzadeh2024relu} change the activation functions in pre-trained models to ReLU and fine-tune to improve activation sparsity. 

Recent work has also addressed the redundancy in the Transformer architecture to achieve reduction of GPU memory usage and speed enhancement \cite{michel2019sixteen,voita-etal-2019-analyzing,ge2024model}. For example, \citet{sajjad2023effect, xia2022structured} investigates the layer redundancy for effective structured pruning. We also highlight input-dependent pruning methods, such as contextual sparsity \citep{liu2023deja} and FastGen \citep{ge2024model}, which should be considered along with the challenges of efficient implementation for optimizing computation and memory. 
Appendix \ref{app:pruning} provides further discussion of pruning techniques.

\subsection{Quantization} \label{sec:model-compression-quant}

Quantization is widely adopted to compress LLMs with vast parameter counts.
The GPTQ~\cite{frantar2022gptq} focuses on layer-wise weight-only quantization, using inverse Hessian matrices to minimize the reconstruction error.
To fully leverage the benefits of fast integer matrix multiplication, more quantization methods~\cite{liu2023qllm, dettmers2022llm, kim2023squeezellm, xiao2023smoothquant, yao2022zeroquant, lin2024awq, liu2023llmqat, liu2024spinquant, liu2023llm, shao2023omniquant} that quantize both weights and activations are increasingly being adopted for LLMs.
AWQ~\cite{lin2024awq} and ZeroQuant~\cite{yao2022zeroquant} take activation into account to assess the importance of weights, enabling more effective optimization for weight quantization. In addition, for K/V Cache Quantization \cite{hooper2024kvquant, liu2024kivi, yue2024wkvquant}, Key-Value cache is specifically quantized for enabling efficient long-sequence length inference.

Another challenge of activation quantization lies in the outliers that fall outside the typical activation distribution.
SmoothQuant~\cite{xiao2023smoothquant} smoothes activation outliers by migrating quantization difficulty from activations to weights.
SpinQuant~\cite{liu2024spinquant} introduces rotation matrices to transform outliers into a new space.
Recently, quantization-aware training (QAT) methods, such as LLM-QAT~\cite{liu2023llmqat} and EdgeQAT~\cite{shen2024edgeqat}, have gained attention due to the strong performance.
Both methods adopt distillation with float16 models to recover the quantizationi error. We also note recent work~\cite{shen2024agile, shen2024edgeqat, zeng2024flightllm} that implements the quantized LLMs on mobile devices and FPGAs to demonstrate the effectiveness and efficiency of the weight and activation quantization for LLMs.

\subsection{Knowledge Distillation Techniques} \label{sec:knowledge-distil-tech}

In its classical form, knowledge distillation~\cite{hinton2015distilling} involves training an efficient model, known as the ``student,'' to replicate the behavior of a larger, more complex model, referred to as the ``teacher.'' In this section, we particularly focus on distillation strategies from one or multiple white-box teacher language model to a target student language model. 

Babyllama~\cite{timiryasov2023babyllamaknowledgedistillation} is among the first to develop a compact 58M parameter language model using a Llama model as the teacher. A key finding of this work is that distillation from a robust teacher can outperform traditional pre-training on the same dataset. In a similar vein, ~\cite{gu2024minillmknowledgedistillationlarge} introduce modifications in the distillation loss, which enables the student models to generate better quality responses with improved calibration and lower exposure bias.  Sequence-level distillation loss can also be improved by using a generalized version of f-divergences as shown in~\cite{wen2023fdivergenceminimizationsequencelevelknowledge}. 
~\citet{liang2023moretaskawarelayerwisedistillation} extend layer-wise distillation strategies for language models by using task-aware filters which distill only the task specific knowledge from the teacher.    
Recent works~\citep{wan2024knowledgefusionlargelanguage, wan2024fusechatknowledgefusionchat} show that multiple language models can be fused as a teacher towards distilling knowledge into small language models by strategically merging their output probability distributions.

One of the issues in knowledge distillation for language models is that the distillation strategies are primarily effective when (1) the teacher and the student language model share the same tokenizer and (2) the teacher's pre-training data is available. ~\citet{boizard2024crosstokenizerdistillationuniversallogit} addresses this issue by introducing an universal logit distillation loss inspired from the optimal transport literature. Often distillation is also combined with pruning techniques towards creating smaller language models. For example, ~\cite{sreenivas2024llmpruningdistillationpractice, muralidharan2024compactlanguagemodelspruning} show that an iterative step of pruning a large language model followed by retraining with distillation losses, can enable strong smaller models. 

Recent advancements have explored methods beyond traditional label distillation by incorporating additional supervision during the distillation process to create smaller language models. ~\citet{hsieh2023distillingstepbystepoutperforminglarger} find that using ``rationales'' as an additional source of supervision during distillation makes it more sample-efficient. Moreover, the authors find that the distilled model outperforms large-language models on commonly used NLI, Commonsense QA and arithmetic reasoning benchmarks. In a similar vein, ~\cite{dai2024imitationlearningkeyreasoning, magister2023teachingsmalllanguagemodels, ho2023largelanguagemodelsreasoning, fu2023specializingsmallerlanguagemodels} distill the reasoning chain from a larger language model to a smaller language model along with the label information. Such distilled models have been shown to possess improved arithmetic, multi-step math, symbolic and commonsense reasoning abilities.

\section{Evaluation} \label{sec:data-and-evaluation}

\begin{table*}
\centering
\resizebox{\linewidth}{!}{
\small   
\begin{tabular}{p{2.5cm} p{2.5cm} p{6.5cm} p{6.5cm}}   
\toprule
\textbf{Setting} & \textbf{Constraints}  & \textbf{Datasets}    & \textbf{Metrics} \\
\midrule
Efficient Inference & Latency      & SuperGLUE \cite{sarlin2020superglue}, SQuAD \cite{rajpurkar2016squad}, TriviaQA \cite{joshi2017triviaqa}, CoQA \cite{reddy2019coqa}, Natural Questions (NQ) \cite{kwiatkowski2019natural} &        Inference Time \cite{narayanan2023cheaply}, Throughput \cite{arora2024simple} \\
\midrule
On-device/Mobile    & Memory       &    TinyBERT \cite{jiao2020tinybert} and OpenOrca \cite{OpenOrca}    & Peak Memory Usage \cite{lee2024designing}, Memory Footprint, Compression Ratio \cite{cao2024retaining} \\

\midrule
Privacy-Preserving  & Privacy       &   PrivacyGLUE \cite{shankar2023privacyglue}, MIMIC \cite{johnson2020mimic}
  
& Privacy Budget \cite{yu2024privacypreserving}, Noise Level \cite{havrilla2024understanding} \\
\midrule
Energy-Efficient AI & Energy Optimization & - & Energy Efficiency Ratio \cite{stojkovic2024dynamollm}, Thermal Efficiency, Idle Power Consumption \cite{patel2024characterizing} \\
\bottomrule
\end{tabular}
}
\caption{Overview of Settings, Constraints, and Metrics.}
\label{table:settings}
\end{table*}

Table \ref{table:settings} presents different evaluation settings along with their corresponding datasets and metrics for SLMs.   
In this section, we examine how different datasets and evaluation metrics are specifically designed to assess SLMs. 
These  evaluation components are organized according to the constraints they address for SLMs.

\subsection{Datasets}
\label{sec:dataset}

The datasets commonly used for pre-training and evaluating SLMs across various settings are outlined in Table \ref{table:settings}. These datasets provide diverse contextual examples that enable models to generalize effectively across different learning settings.

\paragraph{Efficient Inference} This setting requires models to generate output as quickly as possible, with minimal latency and high throughput. Evaluation datasets for this setting often focus on tasks that require fast response times, such as question answering, text classification, and natural language understanding. To this end, some of the example evaluation datasets for this setting can include SuperGLUE \cite{sarlin2020superglue}, SQuAD \cite{rajpurkar2016squad}, TriviaQA \cite{joshi2017triviaqa}, CoQA \cite{reddy2019coqa}, Natural Questions (NQ) \cite{kwiatkowski2019natural}, and many more \cite{chang2024survey} that cover various tasks that require faster response time.

\paragraph{Privacy-preserving} Privacy-preserving datasets play an important role in enabling the development of SLMs while safeguarding sensitive information. Datasets such as PrivacyGLUE \cite{shankar2023privacyglue} apply differential privacy techniques to common tasks such as sentiment analysis. Anonymized datasets such as MIMIC \cite{johnson2020mimic} and n2c2 datasets\footnote{\url{https://portal.dbmi.hms.harvard.edu/projects/n2c2-nlp/}} contain de-identified clinical notes for medical tasks, protecting personal health information. Additionally, federated datasets such as LEAF\footnote{\url{https://github.com/TalwalkarLab/leaf}} allow data to remain distributed across devices, supporting privacy by design through federated learning frameworks.

\paragraph{TinyML and On-device} In these settings, the focus is on deploying SLMs in highly resource-constrained environments. Frameworks such as TinyBERT \cite{jiao2020tinybert} and OpenOrca \cite{OpenOrca} play a pivotal role by enabling the training and evaluation of SLMs on curated datasets tailored for such environments. TinyBERT, a distilled version of BERT, is optimized for both size and speed, making it suitable for on-device applications with minimal latency requirements. Similarly, subsets like OpenOrca provide useful datasets that balance performance and resource constraints, supporting the development of small, efficient models that can be deployed on low-power devices without sacrificing accuracy.

\subsection{Metrics}
The key metrics for evaluating SLMs across different settings are presented in Table \ref{table:settings}. The evaluation metrics are organized based on the specific constraints.

\paragraph{Latency} Two key metrics to evaluate latency are inference time \cite{narayanan2023cheaply} and throughput \cite{arora2024simple}. Inference time measures how quickly a model can process input and generate an output, which is crucial for user-facing applications that require immediate feedback. Throughput, on the other hand, evaluates the number of tokens or samples a model can process in a given period, making it especially relevant for large-scale tasks or time-sensitive applications.

\paragraph{Memory} When deploying models in memory-constrained environments, memory efficiency becomes a primary consideration. Metrics such as peak memory usage \cite{lee2024designing} capture the highest amount of memory the model consumes during inference. Similarly, memory footprint and compression ratio \cite{cao2024retaining} are used to measure how compact a model is and the efficiency of the compression techniques applied, enabling models to operate within memory constraints without sacrificing performance.

\paragraph{Privacy} Privacy budget \cite{yu2024privacypreserving}, a measure rooted in differential privacy, quantifies the model’s ability to protect sensitive information during both training and inference. Alongside this, noise level \cite{havrilla2024understanding} measures the trade-off between privacy and accuracy by assessing how much noise is added to ensure privacy while maintaining the model's performance.

\paragraph{Energy Optimization} The energy efficiency ratio \cite{stojkovic2024dynamollm} evaluates the energy used relative to the model’s overall performance, providing insights into how energy-intensive an SLM is in practice. Other metrics, such as thermal efficiency and idle power consumption \cite{patel2024characterizing}, measure the energy consumed when the model is either actively processing tasks or idle, which is crucial for long-term deployment in energy-constrained environments like embedded systems or mobile devices.
  
\definecolor{GoogleGreen}{HTML}{0F9D58}
\definecolor{GoogleBlue}{HTML}{4285F4}
\definecolor{GoogleRed}{HTML}{EA4335}

\begin{table*}[htbp]
\centering
\small
\resizebox{\linewidth}{!}{
\begin{tabular}{c c H p{5cm} ccccc c}
\toprule

\textbf{Category} & \textbf{Application} & \textbf{Description} & \rotatebox{0}{\textbf{Need for SLM Application}} & \rotatebox{90}{\textbf{Inference Runtime}} & \rotatebox{90}{\textbf{Memory}} & \rotatebox{90}{\textbf{Storage Space}} & \rotatebox{90}{\textbf{Latency}} & \rotatebox{90}{\textbf{Comm. Overhead}} \\

\midrule

\multirow{4}{*}{\textcolor{GoogleGreen}{\textbf{Real-Time Interaction}}} 
& \textbf{Chatbots} & Handle frequent queries and basic troubleshooting. & Real-time response needed, lightweight & \checkmark & \checkmark &  & \checkmark & \checkmark \\
\cmidrule(lr){2-9}
& \textbf{Voice Interfaces} & Used in voice assistants and dictation tools. & Low latency required for real-time & \checkmark & \checkmark &  & \checkmark &  \\
\cmidrule(lr){2-9}
& \textbf{Translation} & Basic translation between languages. & Real-time translation with low-resources & \checkmark & \checkmark &  & \checkmark & \checkmark \\
\midrule

\multirow{5}{*}{\textcolor{GoogleBlue}{\textbf{Content Generation}}} 
& \textbf{Text Summarization} & Summarize articles and reports. & Faster inference, minimal resource use & \checkmark & \checkmark & \checkmark & \checkmark &  \\
\cmidrule(lr){2-9}
\multirow{5}{*}{\textcolor{GoogleBlue}{\textbf{\& Processing}}} 
& \textbf{Sentiment Analysis} & Assess customer sentiment across platforms. & Efficient analysis in low-resource envir. & \checkmark & \checkmark & \checkmark & \checkmark &  \\
\cmidrule(lr){2-9}
& \textbf{Text Classification} & Filter emails, classify content. & Low latency, on-the-fly processing & \checkmark & \checkmark & \checkmark & \checkmark &  \\
\cmidrule(lr){2-9}
& \textbf{NLP for Search} & Improves search engine functionality. & Low latency for real-time search & \checkmark & \checkmark &  & \checkmark &  \\
\cmidrule(lr){2-9}
& \textbf{Autocompletion} & Suggest completions in IDEs or text editors. & Fast prediction with low memory & \checkmark & \checkmark & \checkmark & \checkmark &  \\

\bottomrule
\end{tabular}
}
\caption{Taxonomy of Applications of Small Language Models.}
\label{tab:applications-taxonomy}
\end{table*}

\section{Applications}

\label{sec:applications}

In this section, we consider applications of SLMs, that is, specific use-cases like translation and autocompletion.

\subsection{Real-Time Interaction}
GPT-4o, released in May 2024, processes text, vision, and audio input end-to-end and is faster than GPT-4 Turbo~\cite{gpt4opressrelease}. The demonstration involved responses  in the style of human conversation. LLaMA-Omni combine a speech encoder, adaptor, LLM, and streaming decoder to enable real-time interaction with speech input based on LLaMA-3-8B-Instruct~\cite{fang2024llamaomniseamlessspeechinteraction}. Emotionally Omni-present Voice Assistant, or EMOVA, apply LLaMA-3.1-8B as an end-to-end speech model that can generate poems and describe images at the user's request. Google Deepmind's Project Astra uses Gemini to process audio and video information from a smartphone or glasses and respond to respond to queries like mathematics problems and memorize object sequences~\cite{googleprojectastra}.

\subsection{Content Generation and Processing}
LLMR uses LLMs in mixed reality to generate and modify 3D scenes. It combines language models used in several roles - a Scene Analyzer GPT to summarize objects and give further details like color, Skill Library GPT to determine what is required to fufill a user's request, Builder GPT to generate code for the request, and Inspector GPT to evaluate its code~\cite{delatorre2024llmrrealtimepromptinginteractive}. DreamCodeVR assists users in editing an application in the Unity engine through code generation~\cite{giunchiDreamCodeVRDemocratizingBehavior2024,juliani2020unitygeneralplatformintelligent}. This permits users to edit VR applications without requiring extensive programming knowledge.  

\subsection{Edge Inference and Privacy}
On-device LLMs maintain usability even when 
MobileLLM improve on various chat benchmarks and performs comparably with LLaMA-2-7B in API calling~\cite{liu2024mobilellm}. Apple Intelligence applies an ~3B parameter model to perform on-device inference for a broad range of tasks, such as text and notification summarization, image and emoji generation, and code completion for XCode~\cite{gunter2024appleintelligencefoundationlanguage, wwdcappleintelligence}. On-device inference reduces latency as measured by the time to first generated token~\cite{hu2024inferenceinterferencedisaggregatellm,ggerganov}. HuatuoGPT is a domain-adapted LLM for medical dialogue and BioMistral is an LLM tailored for biomedical work~\cite{huatuogpt-2023, labrak2024biomistralcollectionopensourcepretrained}. Applications related to medicine may need to adhere to stringent privacy regulations and represent a promising area for future work. TalkBack with GeminiNano assists blind and low vision people by describing and captioning images and runs on Android devices~\cite{geminiteam2024geminifamilyhighlycapable}. On-device inference makes this technology usable without an internet connection.

Mixture-of-Experts can reduce inference cost by using a gating network to use only a subset of layers during inference time~\cite{shazeer2017outrageouslylargeneuralnetworks}. Google's GLaM uses mixture-of-experts~\cite{pmlr-v162-du22c} but is a 1.2T parameter model. EdgeMoE extend misture-of-experts to edge computing using an Nvidia Jetson TX2 and Raspberry Pi 4B, with the latter device being CPU-only~\cite{sarkar2023edgemoememoryefficientmultitaskvision}. 
Based on experimental findings that most weights contribute little to the final computation, the authors compress weights and predict the relevant experts in advance.

\section{Open Problems}
\label{sec:open-problems-challenges}

In this section, we discuss open problems and highlight important areas for future work. 
Hallucination and bias are a concern shared by SLMs and LLMs (Section~\ref{sec:problems-hallucination} and~\ref{sec:problems-bias}). 
In Section~\ref{sec:problems-efficiency}, we discuss the increased demand of energy efficiency during inference.
Finally, we examine the privacy risks of SLMs in Section~\ref{sec:problems-privacy}.

\subsection{Hallucination}\label{sec:problems-hallucination}

A pervasive problem with LLMs is hallucination, defined as content that is nonsensical or untruthful in relation to certain sources~\cite{openai2024gpt4technicalreport}. ~\citet{openai2024gpt4technicalreport} propose that as users rely more on models, the harm caused by hallucinations may be increased. Hallucination can be classified into two types: factuality and faithfulness (relevance). With hallucination of factuality, the generation is inconsistent with verifiable facts. In faithfulness hallucination, generation lacks relevance to user queries~\cite{huang2023surveyhallucinationlargelanguage}. HallusionBench, a benchmark for image-context reasoning in vision-language models, found that larger sizes reduced hallucinations~\cite{guan2024hallusionbenchadvanceddiagnosticsuite}. Analysis of the AMBER hallucination benchmark find that the type of hallucination varies as parameter count changes in Minigpt-4~\cite{wang2024amberllmfreemultidimensionalbenchmark}. However, find that bias increases with parameter count for the LLaMA series of models~\cite{zhao2023gptbiascomprehensiveframeworkevaluating}. Future work may need to consider not only how total hallucinations change in SLMs, but also the type and severity may be influenced by model size.

\subsection{Biases}\label{sec:problems-bias}

Language models have been found to reproduce biases present in training data~\cite{brown2020languagemodelsfewshotlearners,openai2024gpt4technicalreport,touvron2023llama}. 
\paragraph{Measuring Bias} Methods for measuring bias such as Bias Benchmark for Question Answering (BBQ)~\cite{parrish2022bbqhandbuiltbiasbenchmark}, RealToxicityPrompts~\cite{gehman2020realtoxicitypromptsevaluatingneuraltoxic}, and Crowdsourced Stereotype Pairs benchmark (CrowS-Pairs)~\cite{nangia-etal-2020-crows}.

\paragraph{Influence of Parameter Count} ~\cite{touvron2023llama} find that larger LLaMA models exhibit increased measured bias on RealToxicityPrompts. ~\cite{zhao2023gptbiascomprehensiveframeworkevaluating} replicate this with StereoSet~\cite{nadeem-etal-2021-stereoset} and their metric GPTBIAS, which uses GPT-4 to classify responses as biased or unbiased. For comparable model sizes, LLaMA-2 had less measured bias than the previous generation~\cite{touvron2023llama2}.

\subsection{Inference-time Energy Use}\label{sec:problems-efficiency}
Energy efficiency is a high priority for SLMs, especially when used on battery-powered devices. \citet{husom2024pricepromptingprofilingenergy} find that architecture significantly influences power consumption using the MELODI benchmar.
CPU-only inference was found to be generally less efficient than on GPU and that laptops require more energy for inference. The authors find response token length to be the most effective predictor of energy usage, suggesting that more concise responses can help to extend battery life.~\citet{stojkovic2024greenerllmsbringingenergyefficiency} find that energy usage can be reduced by about 20\  

\subsection{Data Privacy}\label{sec:problems-privacy}
Privacy concerns can be broadly classified into three categories: training data, the system prompt used at inference time, and the user query. Query privacy is especially important in SLMs.

\paragraph{Training Data}~\citet{li2024llmpbeassessingdataprivacy} address training and system prompt leaking. The authors find that the risk of training data leakage increased faster than their measure of utility for the model series Pythia~\cite{biderman2023pythia}. They also find that data towards the end of pre-training is easier to extract, with attention layers as a possible cause.

\paragraph{System Prompt}~\citet{liu2024promptinjectionattackllmintegrated} describe unauthorized retrieval of the system prompt as prompt leaking and use of the prompt for unintended purposes as prompt abuse. 
They give the example of getting a prompt designed to rephrase user queries to generate code, leading to unexpected cost using Pear AI\footnote{https://www.parea.ai}.

\paragraph{Inference-time Data}
Unlike with the leakage of training data and the system prompt, this primarily impacts the end-users of a model.
In June 2024, Apple announced the application of language models to the digital assistant Siri~\cite{wwdcappleintelligence}. 
In the context of digital assistants, SLMs may need to interface with user data like location history or protected health information.
If such data were used to train or protect a model from misuse, users might face externalities.
Existing literature is limited.

\section{Conclusion} \label{sec:conc}
Given the growing importance of SLMs due to their efficiency and applicability across a wide range of devices and environments, this paper has surveyed SLMs including model architectures, training techniques, and model compression techniques for optimizing SLMs. 
We also introduced an intuitive taxonomy of evaluation metrics for SLMs and summarize various settings and applications where they are important. Furthermore, we summarized the training and benchmark datasets that have been used for SLMs.
Finally, we highlighted the fundamental challenges and open problems that remain to be addressed.
We hope this survey serves as a valuable resource for both researchers and practitioners. driving the next advancements in small yet powerful language models.
  
\section{Limitations}\label{sec:limitations}
  
While SLMs present a broad array of benefits, risks and limitations must also be considered.
Hallucination (discussed in Section~\ref{sec:problems-hallucination}) and reinforcement of societal biases (discussed in Section~\ref{sec:problems-bias}) are widely recognized risks of large language models.
While research has been performed to measure and reduce these behaviors, they have yet to be fully mitigated.~\citet{utama-etal-2020-towards} introduce a framework to reduce self-bias without the specific bias known at test time. Such methods may become more effective with general increases in model capability.
However, risks specific to groups from which researchers are not primarily drawn may remain unrecognized.

\bibliography{main}
\bibliographystyle{acl_natbib}

\clearpage

\appendix

\section{Further Discussion on Pruning Techniques}
\label{app:pruning}

For unstructured pruning for SLMs, we further note that Wanda~\cite{sun2023simple} incorporates both weights and activations into consideration during pruning process, and eliminates the need of weight updates. In addition, the n:m pruning strategy \cite{zhou2021learning} brings unstructured pruning to model acceleration by pruning exactly \textit{n} weights out of every \textit{m}, balancing pruning flexibility and computational efficiency for significant speedups. NVIDIA's TensorRT leverages such sparse patterns to optimize memory access and reduce computational loads, accelerating inference on GPUs, particularly hardware like the A100.
Additionally, the n:m sparse pattern can also be applied in edge AI applications on NVIDIA Jetson Nano to enhance power efficiency and optimize model size. Finally, unstructured pruning often results in sparse matrices requiring specialized hardware or algorithms to maximize computational benefits \cite{frantar2023sparsegpt}.

\end{document}

%% file: preamble.tex
\usepackage{xcolor}
\definecolor{thedarkblue}{RGB}{0,0,120} %104} % 180
\definecolor{mydarkblue}{rgb}{0,0.08,0.45} %ICML dark blue
\definecolor{darkblue}{rgb}{0,0.08,180}
\colorlet{TufteRed}{red!80!black}
\definecolor{theblue}{RGB}{0,0,180}
\colorlet{thered}{TufteRed}
      
\usepackage{microtype}
\usepackage{balance}

\usepackage{booktabs}
\usepackage{tabularx}

\usepackage{amsmath,amssymb,amsthm}

\newcommand{\eat}[1]{\ignorespaces}
\usepackage{comment}

\usepackage{tikz}
\usepackage{verbatim}
\usetikzlibrary{arrows}
\usetikzlibrary{shapes,snakes}
\usetikzlibrary{decorations.pathmorphing} % noisy shapes
\usetikzlibrary{fit}					% fitting shapes to coordinates
\usetikzlibrary{backgrounds}	% drawing the background after the foreground

\usepackage{ragged2e}
\usepackage{multirow}
\usepackage{microtype}
\usepackage{balance}
\usepackage{setspace}

\graphicspath{{./}{./graphics/}}
\newcolumntype{H}{>{\setbox0=\hbox\bgroup}c<{\egroup}@{}}

\newcolumntype{R}[1]{>{\RaggedLeft\arraybackslash}} %p{#1}}
\newcolumntype{L}[1]{>{\RaggedRight\arraybackslash}} %p{#1}}

\AtBeginEnvironment{pmatrix}{\setlength{\arraycolsep}{2pt}}

\DeclareMathOperator{\hugeE}{\mbox{\huge\raise-0.3ex\hbox{E}}}
\DeclareMathOperator{\p}{\mathbb{P}}
\DeclareMathOperator{\hugep}{\mbox{\huge\raise-0.3ex\hbox{$\p$}}}